\documentclass[11pt]{article}
\usepackage{acl2012}
\usepackage{times}
\usepackage{latexsym}
\usepackage{amsmath}
\usepackage{multirow}
\usepackage{graphics}\usepackage{graphicx}
\usepackage{url}

\usepackage{xspace}
\usepackage[dvipsnames,usenames]{color} 
\usepackage[numbers,sort&compress]{natbib} %
\usepackage{amssymb}

\newcommand{\ex}[1]{#1}
\newcommand{\distinctiveness}{distinctiveness\xspace}
\newcommand{\distinctive}{distinctive\xspace}
\newcommand{\generality}{generality\xspace}
\newcommand{\Distinctiveness}{Distinctiveness\xspace}

\newcommand{\Generality}{Generality\xspace}

\newcommand{\nonmemorable}{non-memorable\xspace}
\newcommand{\Dg}{+Google\xspace}
\newcommand{\D}{IMDb-only\xspace} 
\newcommand{\commonlanguage}{``common language''\xspace}

\title{You had me at hello: How phrasing affects memorability}

\newcommand{\asep}{\hspace*{.2in}}
\author{Cristian Danescu-Niculescu-Mizil\asep  Justin Cheng \asep  Jon
  Kleinberg \asep
Lillian Lee \\ Department of Computer Science\\ Cornell University
\\  cristian@cs.cornell.edu, jc882@cornell.edu, kleinber@cs.cornell.edu, llee@cs.cornell.edu}

\date{}

\begin{document}
\maketitle
\begin{abstract}


Understanding the ways in which information achieves widespread
public awareness is a research question of significant interest.  We
consider whether, and how, the way in which the information is
phrased --- the choice of words and sentence structure --- can
affect this process.  To this end, we develop an analysis framework
and build a corpus of movie quotes, annotated with memorability
information, in which we are able to control for both the speaker
and the setting of the quotes.  We find that there are significant
differences between memorable and \nonmemorable quotes in several
key dimensions, even after controlling for situational and
contextual factors.  One is {\em lexical distinctiveness}: in
aggregate, memorable quotes use less common word choices, but at the
same time are built upon a scaffolding of common syntactic patterns.
Another is that memorable quotes tend to be more {\em general} in
ways that make them easy to apply in new contexts --- that is, more
portable.  We also show how the concept of ``memorable language''
can be extended across domains.

\smallskip

{\em To appear at ACL 2012; final version}

\end{abstract}


\newcommand{\citex}[1]{\citep{#1}}
\newcommand{\omt}[1]{}
\newcommand{\cdx}[1]{}

\section{Hello. My name is Inigo Montoya.}\label{sec:intro}

Understanding what items will be retained in the public consciousness, 
and why, is a
question of
fundamental
interest in 
many domains, including
marketing, 
politics,
entertainment, and social media;
as we all know, 
many items barely register, whereas others catch on 
and take hold in many people's minds.

An active line of recent computational work has
employed a variety of perspectives on this 
question.
Building on a foundation in the sociology of diffusion
\citex{rogers-diffusion,strang-diffusion}, 
researchers have explored the ways in which network
structure affects the way information
spreads,
with domains of interest including
blogs \citex{adar-blogspace,Gruhl:2004:IDT:988672.988739},
email \citex{wu2004information}, 
on-line commerce \cite{leskovec-ec06j},
and social media 
\citex{
backstrom-kdd06,
Romero:ProceedingsOfThe20ThInternationalConferenceOn:2011,
sun-page-fanning,
Wu:ProceedingsOfWww:2011}.
There has also been recent research addressing temporal aspects
of how different media sources convey information
\citex{Leskovec:ProceedingsOfKdd:2009,schneider2010visualizing,
yang-temporal-variation}
and
ways in which 
 people react differently to information on different topics
\citex{Romero:ProceedingsOfThe20ThInternationalConferenceOn:2011,
tsur-hashtag-content}.

Beyond all these 
factors, however, 
one's everyday experience with these domains suggests
that the way in which a piece of information is expressed ---
the choice of words, the way it is phrased --- might
also
 have 
a fundamental effect 
on the extent to which it takes hold in people's minds.
Concepts that attain wide reach are often carried in messages
such as political slogans, marketing phrases, or aphorisms
whose language seems intuitively to be memorable, ``catchy,''
or otherwise compelling.

Our first challenge in exploring this hypothesis is to develop a notion
of  ``successful''  language that is precise enough to allow for
quantitative evaluation.  
We also face the challenge of devising an evaluation setting that 
separates the phrasing of a message from the 
conditions
in which it was 
delivered --- highly-cited quotes tend to have
been delivered under compelling circumstances
or fit an existing cultural, political, or social narrative,
and potentially
what appeals to us about the quote is really just its invocation
of these 
extra-linguistic contexts.  
Is the form of the language adding an effect 
{\em beyond or independent of} these 
(obviously very crucial)
factors?
To investigate the question,
one needs a way of controlling ---
as much as possible --- for the role that the surrounding context
of the language plays.

\paragraph{The present work (i): Evaluating language-based memorability}
Defining what makes 
an utterance
memorable
is 
subtle, and 
scholars in several domains have written about this question.
There is a rough consensus that an appropriate definition
involves elements of both {\em recognition} --- people should be
able to retain the quote and recognize it when they hear it invoked ---
and {\em production} --- people should be motivated to refer to
it in  relevant situations
\citep{Harris:CienciasPsicologicas:2008}.
One suggested reason for why some memes succeed is 
their
ability to 
provoke
emotions \cite{heath2001emotional}.
Alternatively,
memorable quotes can be good for expressing
the
feelings, mood, or situation of an individual,
a group, or
a culture (the {\em zeitgeist}):
``Certain quotes exquisitely capture the mood or feeling we
wish to communicate to someone.  We hear them
...
and store them
away for future use''
\citep{Fischoff:AnnualConventionOfTheAmericanPsychologicalAssociation:2000}.

None of these observations, however, serve as definitions, and indeed, we believe it desirable to not pre-commit to an abstract definition, but rather to adopt an operational formulation based on external human judgments.
In designing our study, we focus on a domain in which
(i) there
is rich use of language,
some of which has achieved deep cultural penetration;
(ii) 
there already exist a large number of external human judgments --- perhaps implicit, but in a form we can extract; and
(iii) we can control
for the setting in which the text was used.

Specifically, we use the complete scripts of 
roughly 1000 movies, representing diverse genres, eras, and 
levels of popularity, and consider which lines are the most
``memorable''.
To acquire memorability labels, 
for each sentence in each
script,
we determine whether it 
has been listed as a ``memorable quote''
by users of the widely-known IMDb (the Internet Movie Database),  and also
estimate the number of times it appears on the Web.
Both of these serve as 
memorability metrics for our purposes.

When we evaluate 
properties of memorable quotes, we 
compare
them with quotes that are not assessed as memorable, but
were spoken by the same character,
at approximately the same point in the same movie.
This enables us to control in a fairly fine-grained way for
the confounding effects of context discussed above:
we 
can observe
differences that persist even 
after taking into account both
the speaker and the setting.

In a 
pilot validation study,
we find that human subjects
are effective at recognizing the more 
IMDb-memorable of two quotes,
even for movies 
they have 
not seen.
This motivates a search for features intrinsic to the text of quotes
that signal memorability.
In fact,
comments provided by the human subjects 
as part of
the task suggested two basic forms that such textual signals could take:
subjects felt that 
(i) memorable quotes often involve a {\em distinctive} turn of phrase;
and (ii) memorable quotes tend to invoke {\em general} themes that
aren't tied to the specific setting they came from, and hence
can be 
more easily invoked for future 
(out of context) 
uses.
We test both of these principles in our analysis of the data.

\paragraph{The present work (ii): What distinguishes memorable quotes}

Under
the controlled-comparison setting sketched above, we find that
memorable quotes exhibit significant differences
from non-memorable quotes in several fundamental respects,
and these differences in the data reinforce the two main principles
from the human pilot study.
First, we show a concrete sense in which 
memorable quotes are
indeed
 {\em distinctive}:
with respect to lexical language models trained 
on 
the newswire portions of the Brown corpus \citex{brown},
memorable quotes have significantly lower likelihood than 
their non-memorable counterparts.
Interestingly, this distinctiveness takes place at the level of words,
but not at the level of other syntactic features: 
the part-of-speech composition of memorable quotes is in fact
more likely with respect to 
newswire.
Thus, we can think of memorable quotes as consisting, in an
aggregate sense, of unusual word choices built on a scaffolding
of common part-of-speech 
patterns.

We also identify a number of ways in which 
memorable quotes convey greater {\em generality}.
In their patterns of verb tenses, 
personal pronouns, and determiners, 
memorable quotes are structured so as to be more
``free-standing,'' containing fewer markers that indicate
references to nearby text.

Memorable quotes differ in
other interesting aspects as well,
such as 
sound distributions.

Our analysis of memorable movie quotes suggests a
framework by which the memorability of text in a range of
different domains could be investigated.
We provide evidence that such cross-domain properties may hold,
guided by one of our motivating applications in marketing.
In particular, we analyze a corpus of advertising slogans, and
we show that these slogans have significantly greater likelihood
at both the word level and the part-of-speech level with
respect to a language model trained on memorable movie quotes,
compared to a corresponding language model trained on
non-memorable movie quotes.
This suggests that some of the principles 
underlying memorable text have the potential to apply across 
different areas.

\paragraph{Roadmap} \S\ref{sec:pilot} lays the empirical foundations
of our work: the design and creation of our movie-quotes dataset, which we
make publicly available (\S\ref{sec:data}), a pilot study with human
subjects validating IMDb-based memorability labels
(\S\ref{sec:human}), and further study of incorporating search-engine
counts (\S\ref{sec:web}). \S\ref{sec:experiments} details our analysis
and prediction experiments, using both movie-quotes data and, as an
exploration of cross-domain applicability, slogans
data. \S\ref{sec:relwork}
surveys
related work across a
variety of fields. 
\S\ref{sec:conc} 
briefly summarizes and indicates some future directions.


\def\mem{M}
\def\nonmem{N}

\section{I'm ready for my close-up.}
\label{sec:pilot}

\begin{table*}[t]
\small
\begin{center}
\begin{tabular}{|p{1.06in}||p{2.6in}|p{2.3in}|}
\hline
Movie & First Quote & Second Quote
\\ \hline

Jackie Brown & 
\ex{Half a million dollars will always be missed.} &
\ex{I know the type, trust me on this.}
\\ \hline

Star Trek: Nemesis & 
\ex{I think it's time to try some unsafe velocities.} &
\ex{No cold feet, or any other parts of our anatomy.}
\\ \hline

Ordinary People & 
\ex{A little advice about feelings kiddo; don't expect it always to tickle.} &
\ex{I mean there's someone besides your mother you've got to forgive.}%
\\ \hline

\end{tabular}
\end{center}
\vspace{-3mm} \caption{
Three example pairs of movie quotes.
Each pair satisfies our criteria: the two component quotes
are spoken close together in the movie by the same character, have the same length, 
and one is labeled memorable by the IMDb while the other is not.
(Contractions such as ``it's'' count as two words.)
\label{table:task-examples} \vspace{-2mm}
}
\end{table*}

\subsection{Data}\label{sec:data}

To study
the properties of memorable movie quotes, we need
a source of movie lines and a designation of memorability.
Following \cite{Danescu-Niculescu-Mizil+Lee:11a},
we constructed a corpus consisting of all lines from
roughly 1000 movies,
varying 
in genre, era, and popularity;
for each movie, we then extracted the list of quotes from IMDb's
{\em Memorable Quotes}
page corresponding to the movie.\footnote{
This extraction involved some 
edit-distance-based
alignment, since the exact form of
the line in the script can exhibit minor differences from the
version typed into IMDb.
}

A memorable quote in IMDb can appear either as an 
individual sentence
spoken by one character,
or as a multi-sentence line,
or as a block of dialogue involving multiple characters.
In the latter two cases, it can be hard to determine which particular
portion is viewed as memorable (some involve a build-up to a punch line;
others involve the follow-through after a well-phrased opening sentence),
and so we focus in our comparisons on those memorable quotes that appear
as a single 
sentence
rather than a multi-line block.\footnote{
We also ran experiments relaxing the 
single-sentence
assumption, which
allows for
stricter
scene control and a larger dataset but complicates
comparisons 
involving
syntax.  The non-syntax results were in line
with those reported here.}

\begin{figure}[t]
\begin{center}
\includegraphics[width =.3 \textwidth]{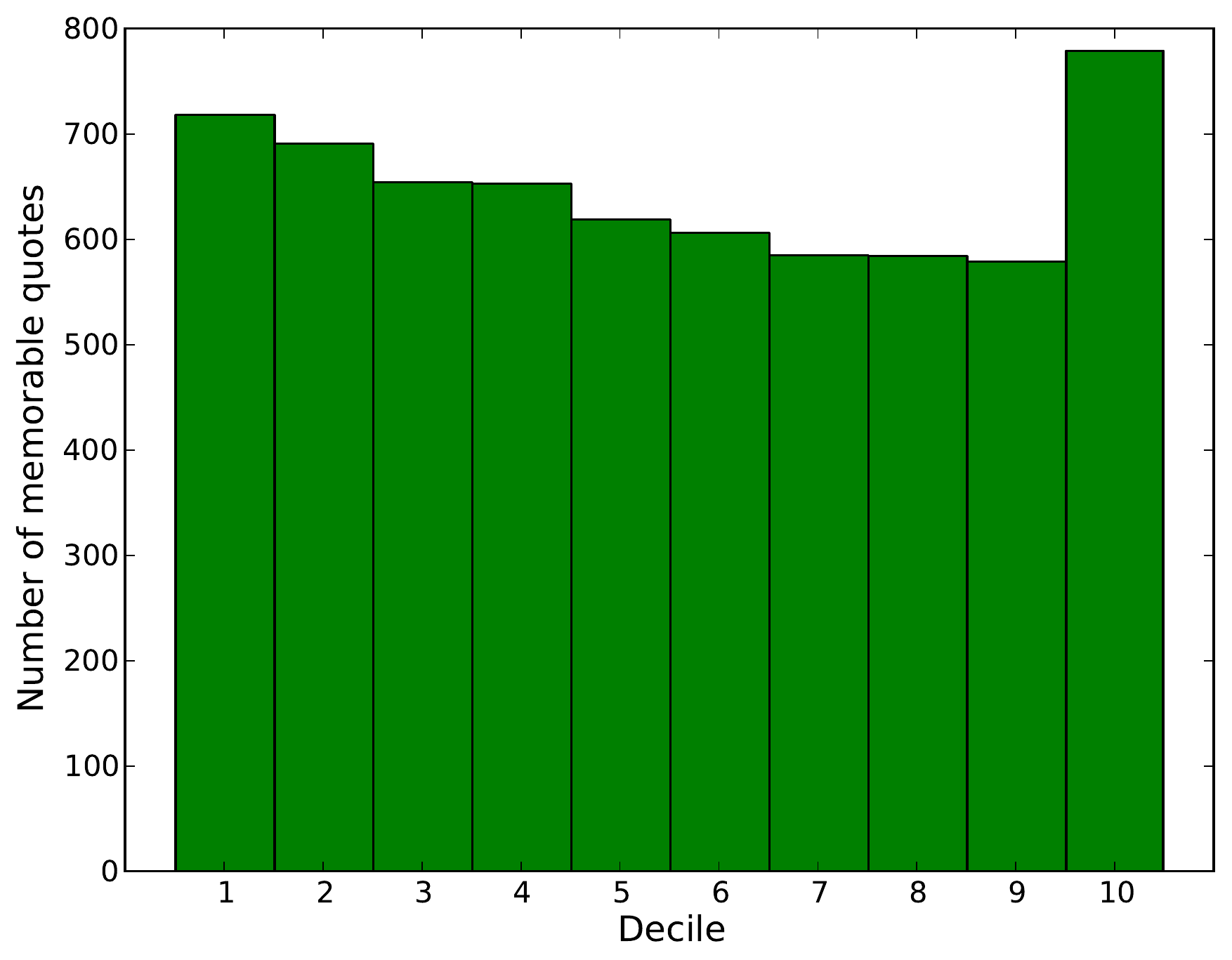}
\end{center}
\caption{
Location of memorable quotes 
in each decile of movie scripts (the first 10th, the second 10th, etc.), summed
over all movies.
The same qualitative results hold if we discard each movie's very first and last line, which might have privileged status.
  \label{fig:deciles}
}
\end{figure}

We now formulate a task that we can use to evaluate the
features of memorable quotes.  Recall that our goal is to identify
effects based in the language of the quotes themselves, beyond 
any factors arising from the speaker or context.
Thus, for each 
(single-sentence)
memorable quote $\mem$, we identify a 
non-memorable quote that is as similar as possible to $\mem$ in all characteristics but the choice of words.  This means we want it to be spoken by the same character in the same movie.  It also means that we want it to have the same length: controlling for length is important because we expect that on average, shorter quotes will be easier
to remember than long quotes, and that wouldn't be an interesting
textual effect to report.  Moreover, we also want to control for the fact that a quote's position in a movie can affect memorability:   certain scenes produce more memorable dialogue, and as 
Figure~\ref{fig:deciles} demonstrates, in aggregate memorable quotes also 
occur disproportionately near the beginnings and especially
the ends of movies.
In summary, then, for each $\mem$, we pick a contrasting (single-sentence) quote $\nonmem$ from the same movie that is as close in the script 
as possible to $\mem$ (either before or after it), subject to the conditions that (i) $\mem$ and $\nonmem$ are uttered by
the same speaker, (ii) $\mem$ and $\nonmem$ have the same number of words, and (iii) $\nonmem$ does not occur in the IMDb list of memorable quotes for the movie (either as a single line or
as part of a larger block).

Given such pairs,
we formulate a 
{\em pairwise comparison task}: given $M$ and $N$,
determine which is the memorable quote.
Psychological research on subjective evaluation
\cite{thurstone1927law}, as well as 
initial experiments using
ourselves
as
subjects,
indicated that this pairwise set-up 
easier
to work with than simply presenting
a single sentence and asking whether it is memorable or not;
the latter requires agreement on an ``absolute'' criterion for memorability
that is very hard to impose consistently, whereas the former simply
requires a judgment that one quote is more memorable than another.

Our main dataset,
available at
\textcolor{blue}{\url{http://www.cs.cornell.edu/~cristian/memorability.html}},\footnote{
Also available there: other examples and factoids.
}
thus consists of 
approximately 
2200
such $(\mem,\nonmem)$ pairs,
separated by a median of 5 same-character lines in the script.  
The reader can get a sense for the nature of the 
data
from 
the three examples in Table \ref{table:task-examples}.  

We now discuss two further aspects to the formulation of the experiment:
a preliminary pilot study involving human subjects, and
the incorporation of search engine counts into the data.

\subsection{Pilot study: Human performance}
\label{sec:human}

{
\begin{table}\small
\begin{center}
\begin{tabular}{|c|r@{  = }r@{\%}r|} \hline
subject  & \multicolumn{3}{c|}{number of matches with} \\ & \multicolumn{3}{c|}{IMDb-induced
  annotation} \\ \hline
A & 11/11 & 100 & \\
B &    11/12  & 92 & \\
C &  9/11  & 82 & \\
D  &  8/11  & 73 & \\ 
E   &  7/11  & 64 & \\
F   & 7/12  & 58 & \\
\hline macro avg & \multicolumn{1}{c}{---}  &  78 & \\\hline
\end{tabular}
\end{center}
\caption{\label{tab:task-perf} Human pilot study: number of matches to 
IMDb-induced
  annotation, ordered by decreasing match percentage.  
For the null hypothesis 
of random guessing, 
these results are statistically
  significant, $p < 2^{-6} \approx .016$.
}
\end{table}
}

As a preliminary consideration,
we did a small pilot study
to 
see if humans can distinguish
memorable from non-memorable quotes, assuming our IMDB-induced labels
as gold standard.  Six subjects, all native
speakers of English and none an author of this paper, were presented with 11 or 12 pairs of memorable
vs. non-memorable quotes;
again, we controlled for extra-textual effects by ensuring that in each pair the two quotes come from the same movie,  are by the same character, have the same
length, and appear as nearly as possible in the same scene.\footnote{
In this pilot study, we 
allowed multi-sentence quotes.
}
 The order
of quotes within pairs was randomized.
Importantly, because we wanted to understand whether the language of the quotes 
by itself contains signals about memorability, we chose quotes from
movies that the subjects said they had not seen. (This means 
that each subject saw a different set of quotes.)  Moreover, the
subjects were requested not to consult any external sources of
information.\footnote{We did not use crowd-sourcing
because we saw no way to ensure
 that this condition would be
  obeyed by arbitrary subjects.
We do note, though, that after our research was completed and as of
Apr. 26, 2012,  $\approx$
11,300 people completed the online test: average accuracy: 72\%, mode
number correct: 9/12.}
The reader is welcome to try a demo version of the task at \textcolor{blue}{\url{http://www.cs.cornell.edu/~cristian/memorability.html}}.

Table \ref{tab:task-perf} shows that
all the subjects
performed (sometimes much) better than chance, and against the null hypothesis that all subjects are guessing randomly,
the results are statistically significant, $p < 2^{-6} \approx .016$.
These preliminary findings provide evidence for
the validity of 
our task:
despite the apparent difficulty of the job,
even humans who haven't seen the
movie in question  can recover
our IMDb-induced labels
with some reliability.\footnote{
The average accuracy being below 100\% reinforces that context 
is  
very
important, too.
}

\subsection{Incorporating search engine counts}
\label{sec:web} 

Thus far we have discussed a dataset in which memorability is determined
through an explicit labeling drawn from the IMDb.
Given
the ``production'' aspect of
memorability discussed in 
\S\ref{sec:intro},
we should also expect that memorable quotes will tend to appear
more extensively on Web pages than non-memorable quotes;
note that incorporating this insight makes it possible to use the
(implicit) judgments of a much larger number of people than are represented by the IMDb database.
It therefore makes sense to try using 
search-engine result
counts
as a second indication of memorability.

We experimented with several ways of constructing memorability information from search-engine counts, but this proved challenging.
Searching for a quote as a stand-alone phrase
runs into the problem
that a number of 
quotes are also sentences that people use
without the movie in mind, and so high counts for such quotes do not
testify to the phrase's status as a memorable quote from the movie.
On the other hand, searching for the quote in a
Boolean conjunction with the 
movie's title
discards
most of these uses, but also eliminates a large
fraction of the appearances on the Web that we want to find: 
precisely because memorable quotes tend to have widespread cultural usage,
people
generally don't feel the need to include the
movie's title when invoking them.
Finally, since we are dealing with roughly 1000 movies, the
result counts vary over an enormous range, from recent blockbusters
to movies with relatively small fan bases. 

In the end, we found that it was more effective to use the result counts
in conjunction with the IMDb labels, so that the counts 
played
the role of an additional filter rather than a free-standing numerical value.
Thus, for each pair $(M,N)$ produced using the IMDb methodology above, 
we searched for each of $M$ and $N$ as quoted expressions in a Boolean
conjunction with the title of the movie.  We then kept only those pairs
for which $M$ (i) produced 
more than
 five results in our 
(quoted, conjoined) search, and (ii) produced at least twice as many
results as the corresponding search for $N$.
We
created
a version of this filtered dataset using each of Google and Bing,
and all the main findings were consistent with the results on the
IMDb-only dataset.
Thus, in
 what follows, we will focus on the main IMDb-only dataset,
discussing the relationship to the dataset filtered by search engine counts
where relevant
(in which case we will refer to the \Dg dataset).


\section{\mbox{Never send a human to do a machine's job.}}
\label{sec:experiments}

We now discuss
experiments that investigate the  hypotheses
discussed in 
\S\ref{sec:intro}. 
In particular, we devise
methods that can assess the 
\distinctiveness and \generality hypotheses
and
test
whether
 there exists a notion of ``memorable language''
that operates across domains.  In addition, we evaluate and
compare the predictive power of these hypotheses.

\subsection{Distinctiveness}

One of the  hypotheses we examine is whether the use of language
in memorable quotes  is to some extent unusual.   In order to quantify
the level of \distinctiveness  of a quote, we take a language-model 
approach: we model ``common language'' using 
the newswire sections of the Brown corpus \cite{brown}\footnote{Results were qualitatively similar if we used the fiction portions.  The age of the Brown corpus makes it less likely to contain  modern movie quotes.},
and evaluate how
\distinctive a quote is by evaluating its likelihood with respect to
this model --- the lower the likelihood,
the more \distinctive.
In order to assess different 
levels
of
lexical and 
syntactic \distinctiveness, we employ a total of
six 
Laplace-smoothed\footnote{We employ Laplace (additive) smoothing with
  a smoothing parameter of 0.2.
The language
 models' vocabulary was that of the entire training corpus.}
language models:  1-gram, 2-gram, and 3-gram word LMs and 1-gram,
2-gram and 3-gram part-of-speech\footnote{Throughout we obtain part-of-speech tags by using the NLTK maximum entropy tagger 
with 
default parameters.}
 LMs.

We find strong evidence that from a lexical perspective, memorable
quotes are more \distinctive than their \nonmemorable counterparts. As
indicated in  Table \ref{table:distinctiveness},
for each of our lexical ``common language'' models, 
in 
about
60\% of the
quote pairs, the memorable quote is  more \distinctive.

 Interestingly, the reverse is
true when it comes to syntax: memorable quotes
appear to follow the syntactic patterns of \commonlanguage as
closely as 
 or more
closely than \nonmemorable quotes.
  Together, these results suggest
that memorable quotes consist of unusual word sequences built on
common syntactic scaffolding.  

\begin{table}[t]
\begin{center}
{\small
\begin{tabular}{|cc|l|l|}
\hline
\multicolumn{2}{|c|}{\begin{tabular}{c}\commonlanguage \\model\end{tabular}}  & \D  & \Dg
\\ \hline
\multirow{3}*{lexical}  & 1-gram  & $61.13\%^{***}$ &$59.21\%^{***}$\\
&2-gram   & $59.22\%^{***}$&$57.03\%^{***}$\\
&3-gram  & $59.81\%^{***}$&$58.32\%^{***}$\\
\hline
\multirow{3}*{syntactic} & 1-gram  & $43.60\%^{***}$ &$44.77\%^{***}$\\
&2-gram   & $48.31\%$&$47.84\%$\\
&3-gram  & $50.91\%$&$50.92\%$\\
\hline
\end{tabular}
}
\end{center}
\vspace{-3mm} \caption{\Distinctiveness: percentage of quote pairs in
  which the the memorable quote is more distinctive than the
  \nonmemorable one according to the respective \commonlanguage
  model. Significance according to a two-tailed sign test is indicated
  using *-notation ($^{***}$=``p$<$.001''). 
\label{table:distinctiveness} 
}
\end{table}

\subsection{\Generality}
Another of our  hypotheses is that memorable quotes are easier to use outside the specific context in which they were uttered
--- that is, more ``portable'' ---
and therefore 
exhibit fewer terms that refer to those settings. 
We use the following syntactic properties as proxies for the \generality of a quote:
\begin{itemize}
\item 
\textbf{Fewer 3\textsuperscript{rd}-person pronouns}, 
since these commonly refer to a person
or object
 that was introduced earlier in the discourse.   Utterances that employ fewer 
such
pronouns are easier to adapt to new contexts, and so will  be  considered more general.
\item\textbf{More indefinite articles} like \textit{a} and \textit{an}, since they  are
  more likely to refer to general concepts than 
definite articles.
Quotes with
more indefinite articles will be considered more general.
\item \textbf{Fewer past tense verbs and more present tense verbs}, since the former are more likely to refer to specific previous events.
Therefore utterances that employ fewer past tense verbs (and more present tense verbs)
will be considered
more general.
\end{itemize}

Table \ref{table:generality} gives the results for each of these
four metrics --- in each case, we show the percentage of quote pairs
for which the memorable quote scores better on the generality metric.

Note that because the issue of generality is a complex one for which
there is no straightforward single metric, our approach here is based on
several proxies for generality, considered independently; yet, as the results show,
all of these point in a consistent direction.  
It is an interesting open question 
to develop richer ways of assessing whether
a quote has greater generality, in the sense that people intuitively
attribute to memorable quotes.

\begin{table}[t]
\begin{center}
{\small
\begin{tabular}{|l|l|l|}
\hline
\Generality metric  & \D  & \Dg \\
\hline
fewer 3\textsuperscript{rd} pers. pronouns &$64.37\%^{***}$&$62.93\%^{***}$\\
more indef. article & $57.21\%^{***}$&$58.23\%^{***}$\\
less past tense & $57.91\%^{***}$&$59.74\%^{***}$\\
more present tense & $54.60\%^{***}$&$55.86\%^{***}$\\
\hline
\end{tabular}
}
\end{center}
\vspace{-3mm} \caption{\Generality: percentage of quote pairs in which the memorable quote is more general than the \nonmemorable ones according to the respective metric. Pairs where the metric does not distinguish between the quotes are not considered.
\label{table:generality} 
}
\end{table}

\subsection{``Memorable'' language beyond movies}

One of the motivating questions in our analysis is whether
there are general principles underlying ``memorable language.''
The results thus far suggest potential families of such principles.
A further question in this direction is whether the notion of
memorability can be extended across different domains, and for this we 
collected (and distribute on our website)
 431 phrases that were explicitly designed to be memorable: advertising slogans (e.g., ``Quality never goes out of style.'').
The focus on slogans is also in keeping with one of the 
initial motivations in studying memorability, namely, marketing applications ---
in other words, assessing whether a proposed slogan has features
that are consistent with memorable text.

The fact that it's not clear how to construct a collection of ``\nonmemorable'' counterparts
to slogans appears to pose a technical challenge.  However, we can still use a language-modeling approach
to assess whether the textual properties of the slogans are closer to the 
memorable movie quotes (as one would conjecture) or to the 
\nonmemorable movie quotes.
Specifically, 
we train one language model on memorable quotes and
another on \nonmemorable quotes and compare how likely each slogan is to
be produced according to these two models.  As shown in
the middle column of
Table \ref{table:slogans}, we find that  slogans are better predicted both
lexically and syntactically by the former model. 
This result thus offers evidence for a concept of ``memorable language''
that can be applied beyond a single domain.

We also note that the higher likelihood of slogans under a ``memorable language'' model is not simply occurring 
for the trivial reason that this model  predicts all other large bodies of text better.
In particular, the 
newswire section of the
Brown corpus is predicted better
at the lexical level by the language model trained
on \nonmemorable quotes.

\begin{table}[t]
\begin{center}
{\small
\begin{tabular}{|cc|l|l|}
\hline
\multicolumn{2}{|c|}{\begin{tabular}{c}(Non)memorable\\language models\end{tabular}}  & Slogans & Newswire
\\ \hline
\multirow{3}*{lexical}  & 1-gram  & $56.15\%^{**}$ &  $33.77\%^{***}$\\
&2-gram   & $51.51$\%&$25.15\%^{***}$\\
&3-gram  & $52.44\%$&$28.89\%^{***}$\\
\hline
\multirow{3}*{syntactic}  & 1-gram  & $73.09\%^{***}$ &$68.27\%^{***}$\\
&2-gram   & $64.04\%^{***}$&$50.21\%$\\
&3-gram  & $62.88\%^{***}$&$55.09\%^{***}$\\
\hline
\end{tabular}
}
\end{center}
\vspace{-3mm} \caption{Cross-domain concept of ``memorable'' language: percentage of slogans that have higher likelihood under the memorable language model than under the \nonmemorable one (for each of the six language models considered).  Rightmost column: for reference, the percentage of newswire sentences that have higher likelihood under the memorable language model than under the \nonmemorable one.
\label{table:slogans} 
}
\end{table}

Finally, Table \ref{table:generality2} shows that slogans employ
general language, in the sense that for each of our  generality metrics, we see a 
slogans/memorable-quotes/non-memorable quotes spectrum.
\begin{table}[t]
\begin{center}
{\small
\begin{tabular}{|l|r|r|r|}
\hline
\Generality metric  & slogans  & mem. & n-mem. \\
\hline
\% 3\textsuperscript{rd} pers. pronouns &$2.14\%$&$2.16\%$&$3.41\%$\\
 \% indefinite articles & $2.68\%$&$2.63\%$&$2.06\%$\\
\% past
tense
& $14.60\%$&$21.13\%$&$26.69\%$\\
\hline
\end{tabular}
}
\end{center}
\vspace{-3mm} \caption{Slogans are most general when compared to
  memorable and non-memorable quotes.  
  (\%s of 
  3\textsuperscript{rd} pers.
  pronouns and indefinite articles are relative to all tokens, \%s of past tense are relative to all past and present verbs.)
\label{table:generality2} 
}
\end{table}

\subsection{Prediction task}
\label{sec:predict}

We now show how the principles discussed above can 
provide features for a basic prediction task, corresponding to 
the 
task in our human pilot study: given a pair of quotes, identify
the memorable one.

Our first formulation of the prediction task uses a 
standard
bag-of-words model\footnote{We discarded terms appearing fewer than 10 times.}.  If there were no information in the textual content of a quote to determine whether it were memorable, then an SVM employing bag-of-words features should perform no better than chance.  Instead, though, it  obtains 59.67\% 
(10-fold cross-validation) 
accuracy,
as shown in Table \ref{table:prediction}.
We then develop models using features based on the measures formulated earlier in this section: 
generality measures 
(the four listed in Table \ref{table:generality});
distinctiveness measures (likelihood according to 1, 2, and 3-gram \commonlanguage models at the lexical and part-of-speech level for each quote in the pair, their 
differences, and pairwise comparisons between them); and similarity-to-slogans measures  (likelihood according to 1, 2, and 3-gram slogan-language models at the lexical and part-of-speech level for each quote in the pair, their 
differences, and pairwise comparisons between them).

Even a relatively small number of distinctiveness features,
on their own, improve significantly over the much larger bag-of-words model.
When we include additional 
features based on generality
and language-model features measuring similarity to slogans,
the performance improves further (last line of Table \ref{table:prediction}).

Thus, the main conclusion from these prediction tasks is that abstracting
notions such as distinctiveness and generality can produce relatively
streamlined models that outperform much heavier-weight bag-of-words models,
and can suggest steps toward approaching the performance of human judges
who --- very much unlike our system --- have the full cultural context in which movies occur at their disposal.

\begin{table}[t]
\begin{center}
{\small
\begin{tabular}{|l|r|l|}
\hline
Feature set  & \multicolumn{1}{c|}{\# feats} & \multicolumn{1}{c|}{Accuracy}
\\ \hline
bag of words & $962$ & $59.67\%$  \\
\hline
\distinctiveness & $24$ & $62.05\%^{*}$ \\
\generality & 4 & $56.70\%$  \\
slogan sim. & $24$ & $58.30\%$  \\
{\it all three types together} & $52$ & $64.27\%^{**}$ \\
\hline
\end{tabular}
}
\end{center}
\vspace{-3mm} \caption{Prediction: SVM 10-fold cross validation results using the respective  feature sets. Random baseline accuracy is 50\%.  Accuracies statistically significantly greater than bag-of-words according to a two-tailed t-test are indicated with *(p$<$.05) and **(p$<$.01).
\label{table:prediction} 
}
\end{table}

\subsection{Other characteristics}

We also made some auxiliary observations that
may be of interest.
Specifically,
we find differences in letter and sound distribution (e.g., memorable
quotes
--- after curse-word removal ---
 use significantly more
``front sounds'' (labials or front vowels such as represented by the letter \textit{i})
and significantly fewer
``back sounds'' such as
the one represented by \textit{u}),\footnote{%
These findings may relate to marketing research on {\em sound symbolism} \citep{Klink:MarketingLetters:2000,Yorkston:JournalOfConsumerResearch:2004,Colapinto:2011a}.} 
word complexity  (e.g., memorable quotes use words with significantly  more syllables) and phrase complexity (e.g., memorable quotes use fewer coordinating conjunctions).  The latter two are in line with our distinctiveness hypothesis.


\section{A long time ago, in a galaxy far, far away}
\label{sec:relwork} 

How an item's linguistic form affects the reaction it generates has been studied in several contexts, including evaluations of product reviews \cite{DanescuNiculescuMizil+al:09a}, 
political speeches
\citep{guerini2008trusting},
on-line posts 
\citep{Guerini:ProceedingsOfIcwsm:2011},
scientific papers \citep{Guerini:ProceedingsOfIcwsm:2012},
and retweeting of Twitter posts
\citep{tsur-hashtag-content}.
We use a different set of features,
abstracting the notions of distinctiveness and generality, 
in order to focus on these higher-level aspects of phrasing rather
than on particular lower-level features.

Related to our interest in distinctiveness, work in 
advertising research has studied the effect of syntactic
complexity on recognition and recall of slogans
\citep{Bradley:PsychologyAndMarketing:2002,
Lowrey:JournalOfAdvertising:2006,Chamblee:JournalOfAdvertisingResearch:1993}.
There may also be connections 
to
Von Restorff's {\em isolation effect} 
\citet{Hunt:PsychonomicBulletinReview:1995},
which asserts that
when all
but one item in a list are similar in some way, memory for
the different item is enhanced.

Related to our interest in generality, 
\citet{knapp1981memorable} surveyed subjects regarding memorable
messages or pieces of advice they had received, finding
that the 
ability to be applied to 
multiple
concrete situations was an
important factor.

Memorability, although distinct from ``memorizability'', 
relates to 
short-
and
long-term recall.
\citet{Thorn:InteractionsBetweenShortTermAndLongTermMemoryIn:2009}
survey sub-lexical, lexical, and semantic
attributes affecting short-term memorability of lexical items.
Studies of verbatim recall have also considered the 
task of distinguishing an exact quote from close paraphrases
\citep{Bates:JournalOfExperimentalPsychologyHumanLearningAnd:1980}.
Investigations of long-term recall have included studies of
culturally significant passages of text 
\citep{Rubin:JournalOfVerbalLearningAndVerbalBehavior:1977}
and findings 
regarding the effect 
of 
rhetorical devices of
alliterative \citep{Boers:System:2005}, ``rhythmic, poetic, and thematic constraints'' \citep{HymanJr.:MemoryCognition:1990,Parry:TheMakingOfHomericVerseTheCollected:1971}.

Finally,
 there are complex connections between humor and memory
\citep{Summerfelt:TheJournalOfGeneralPsychology:2010},
which may lead to interactions with  computational humor recognition
\citep{Mihaelcea+Strapparava:2006a}.


\section{I think this is the beginning of a beautiful friendship.} 
\label{sec:conc}

Motivated by the broad question of what kinds of information
achieve widespread public awareness, we studied the 
the effect of phrasing on a quote's memorability.
A challenge is that quotes differ not only
in how they are worded, but also in who said them and under
what circumstances; to deal with this difficulty, we constructed a
controlled corpus of movie 
quotes in which lines
deemed memorable are paired with \nonmemorable lines spoken 
by the same character at approximately the same point in the same movie.
After controlling for context and situation, 
memorable quotes were
still
 found to exhibit,
{\em  on average} (there will always be individual exceptions),
significant differences from \nonmemorable quotes in several
important respects, including measures capturing
distinctiveness and generality.
Our experiments with slogans show how 
the principles we identify can extend to a different domain.

Future work
may lead to applications in marketing, advertising and education \citep{Boers:System:2005}.
Moreover, the subtle nature of memorability, and its connection to
research in psychology, suggests a range of 
further research directions.
 We believe 
that the framework developed
here
can serve as the basis for further computational studies of the
process by which information takes hold in the public consciousness,
and the role that language effects play in this process.


{\small
\paragraph*{My mother thanks you. My father thanks you. My sister thanks you. And I thank you:}
\newcommand{\fn}[2]{#1#2}
\fn{R}ebecca Hwa, 
\fn{E}{vie} Kleinberg, 
\fn{D}{iana} Minculescu, %
\fn{A}{lex} Niculescu-Mizil,
\fn{J}{ennifer} Smith, 
\fn{B}enjamin Zimmer, 
and the anonymous reviewers
for helpful discussions and comments; our annotators
\fn{S}{teven} An, \fn{L}{ars} Backstrom, \fn{E}{ric} Baumer,
\fn{J}{eff} Chadwick, \fn{E}{vie} Kleinberg, and \fn{M}{yle} Ott; 
and the makers of Cepacol, Robitussin, and Sudafed, whose
products
got us through the submission deadline.
This paper is based upon work supported in part by 
NSF grants IIS-0910664, IIS-1016099, Google, and Yahoo!
}

\newcommand{\bibsnip}{}

\end{document}